\documentclass[letterpaper, 10 pt, conference]{ieeeconf}  
\makeatletter
\let\NAT@parse\undefined
\makeatother
\IEEEoverridecommandlockouts                              
\usepackage{graphicx} 
\usepackage[numbers,sort&compress]{natbib}  
\usepackage{float}  
\usepackage{comment}
\usepackage{booktabs}
\usepackage{amsmath}
\usepackage{hyperref}
\usepackage{amssymb}

\usepackage{acro}
\DeclareAcronym{RRT}{short = RRT, long = Rapidly-exploring Random Trees}
\DeclareAcronym{DWA}{short = DWA, long = Dynamic Window Approach}
\DeclareAcronym{PRM}{short = PRM, long = Probabilistic Road Maps}
\DeclareAcronym{RVO}{short = RVO, long = reciprocal velocity obstacles, class = abbrev}
\DeclareAcronym{VO}{short = VO, long = velocity obstacles, class = abbrev}
\DeclareAcronym{IGP}{short = IGP, long = interacting Gaussian processes, class = abbrev}
\DeclareAcronym{ANN}{short = ANN, long = artificial neural network, class = abbrev}
\DeclareAcronym{RNN}{short = RNN, long = recurrent neural network, class = abbrev}
\DeclareAcronym{RL}{short = RL, long = reinforcement learning, class = abbrev}
\DeclareAcronym{MPC}{short = MPC, long = Model Predictive Control, class = abbrev}
\DeclareAcronym{UAV}{short = UAV, long = unmanned aerial vehicle, class = abbrev}
\DeclareAcronym{ME-IRL}{short = ME-IRL, long = Maximum Entropy Inverse Reinforcement Learning}
\DeclareAcronym{IROS}{short = IROS, long = International Conference on Intelligent Robots and Systems}
\DeclareAcronym{ICRA}{short = ICRA, long = International Conference on Robotics and Automation}
\DeclareAcronym{RSS}{short = RSS, long = Robotics: Science and Systems}
\DeclareAcronym{IJRR}{short = IJRR, long = International Journal of Robotics Research}
\DeclareAcronym{JFR}{short = JFR, long = Journal of Field Robotics}
\DeclareAcronym{TEB}{short = TEB, long = timed elastic band}
\DeclareAcronym{POMDP}{short = POMDP, long = partially observable Markov decision process}
\DeclareAcronym{CADRL}{short = CADRL, long = Collision Avoidance with Deep Reinforcement Learning}
\DeclareAcronym{MCTS}{short = MCTS, long = Monte Carlo Tree Search}
\DeclareAcronym{SLAM}{short = SLAM, long = Simultaneous Localization and Mapping}

\DeclareAcronym{MDP}{short = MDP, long = Markov decision process}
\DeclareAcronym{PPO}{short = PPO, long = proximal policy optimization}
\DeclareAcronym{LSTM}{short = LSTM, long = long short-term memory network}
\DeclareAcronym{VAE}{short = VAE, long = variational auto-encoder}
\DeclareAcronym{GAN}{short = GAN, long = generative adversarial network}
\DeclareAcronym{CNN}{short = CNN, long = convolutional neural network}
\DeclareAcronym{SARL}{short = SARL, long = Socially Attentive Reinforcement Learning}  
\DeclareAcronym{FCN}{short = FCN, long = fully connected network}

\title{\LARGE \bf
NavRep: Unsupervised Representations for Reinforcement Learning of Robot Navigation in Dynamic Human Environments
}

\author{Daniel Dugas, Juan Nieto, Roland Siegwart and Jen Jen Chung
\thanks{This work was supported by the EU H2020 project CROWDBOT under grant nr. 779942}
\thanks{The authors are with the Autonomous Systems Lab, ETH Z{\" u}rich, Z{\"u}rich 8092, Switzerland. {\tt\small\{dugasd; jnieto; rsiegwart; chungj\}@ethz.ch}}%
}

\begin{document}


\maketitle
\thispagestyle{empty}
\pagestyle{empty}

\begin{abstract}
Robot navigation is a task where reinforcement learning approaches are still unable to compete with traditional path planning.
State-of-the-art methods differ in small ways, and do not all provide reproducible, openly available implementations. This makes comparing methods a challenge.
Recent research has shown that unsupervised learning methods can scale impressively, and be leveraged to solve difficult problems.
%
In this work, we design ways in which unsupervised learning can be used to assist reinforcement learning for robot navigation.
We train two end-to-end, and 18 unsupervised-learning-based architectures, and compare them, along with existing approaches, in unseen test cases. We demonstrate our approach working on a real life robot.
%
%
Our results show that unsupervised learning methods are competitive with end-to-end methods. We also highlight the importance of various components such as input representation, predictive unsupervised learning, and latent features.
We make all our models publicly available, as well as training and testing environments, and tools\footnote{\url{https://github.com/ethz-asl/navrep}}.
This release also includes OpenAI-gym-compatible environments designed to emulate the training conditions described by other papers, with as much fidelity as possible.
Our hope is that this helps in bringing together the field of RL for robot navigation, and allows meaningful comparisons across state-of-the-art methods.
\end{abstract}

\section{Introduction}\label{sec:introduction}

\begin{figure}[t]
    \centering
    \includegraphics[width=\linewidth]{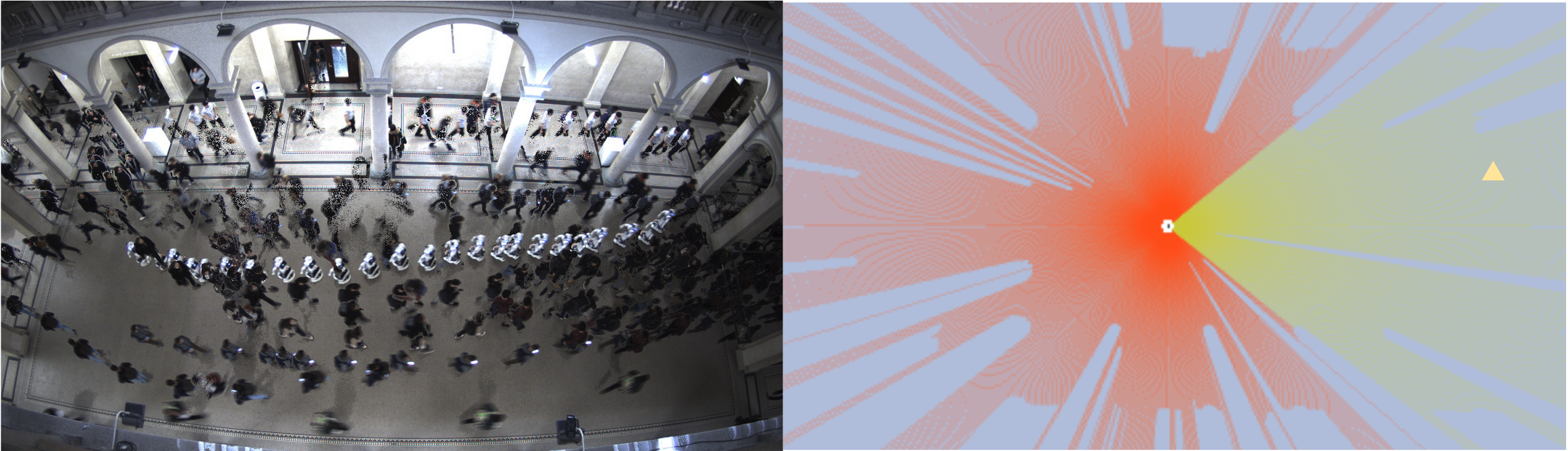}
	\caption{Timelapse of an example target environment for autonomous navigation, snapshot of the corresponding LiDAR measurements on the right}
	\label{fig:domain}
\end{figure}

Robot navigation in complex environments and in the presence of humans is a challenging problem due to complexity in human behavior 
and the unpredictability of unstructured environments (see Fig.~\ref{fig:domain}).
Though popular planning approaches work well in many scenarios, edge cases complex enough to cause planning failures abound, and the engineering work which is necessary to address one edge case does not necessarily translate to another.
As a result, designing a planner that is able to perform in any environment is still a monumental engineering task.

Outside of robot navigation, \ac{RL} approaches have led data-driven controllers to reach expert and even super-human performance at game-play tasks \cite{badia2020agent57}. 
Data-driven methods, at least in theory, present a potential of unlimited scale, where a planner can be improved indefinitely so long as more data can be gathered.
For this reason, several research efforts have focused on the use of data driven-methods for robot navigation~\cite{pfeiffer2017perception,pfeiffer2018reinforced,fan2020distributed}.
Yet, the question remains of whether similar performance can be achieved for the problem of navigation among humans, and how.

\begin{figure}[t]
    \centering
    \includegraphics[width=\linewidth]{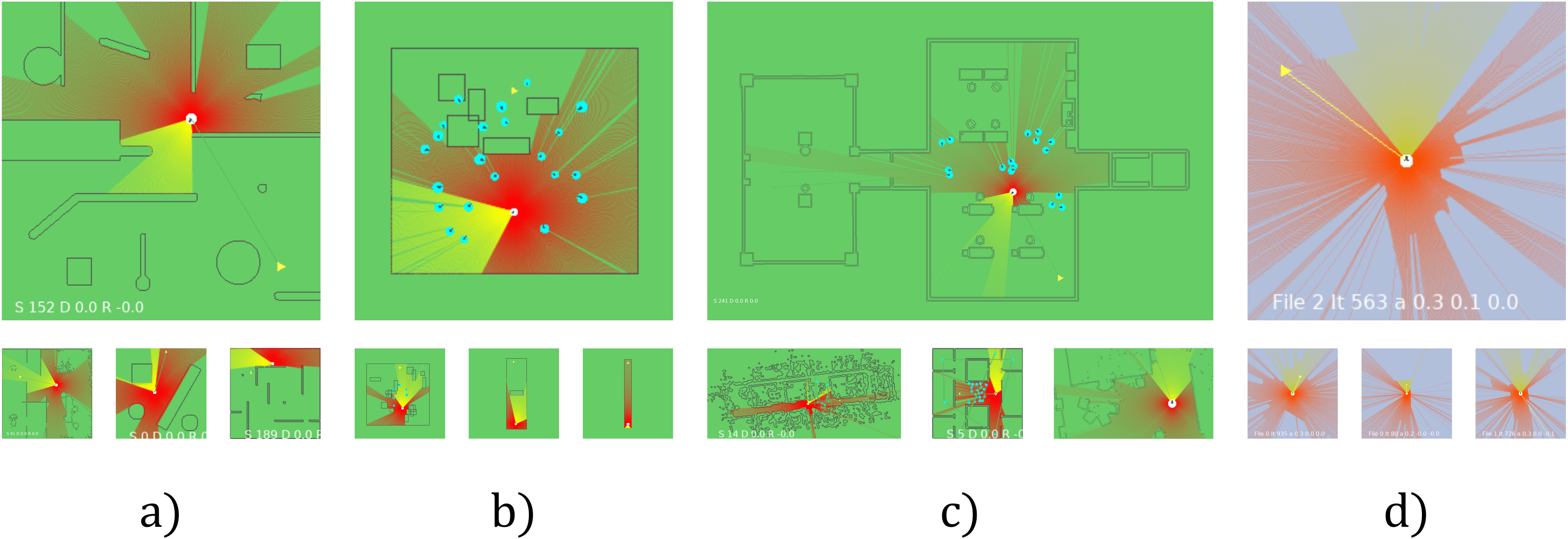}
	\caption{We make NavRepSim openly available, a simulator which contains efficient, gym-compatible environments for end-to-end navigation, allowing all to reproduce scenes from (a)~\cite{pfeiffer2017perception, pfeiffer2018reinforced}, (b) CADRL and SOADRL~\cite{chen2017decentralized, chen2019crowd, liu2020soadrl}, CrowdMove \cite{fan2020distributed} (not shown), (c) IAN~\cite{dugas2020ian}, and (d) real data. } 
	\label{fig:envs}
\end{figure}

When planning in the presence of humans, it is often necessary to provide estimates for the position and state of humans around the robot~\cite{chen2017decentralized,chen2019crowd,liu2020soadrl,chen2017socially,everett2018motion}. To make planning tractable, these estimates are typically reduced to position, velocity, and radius of the whole human.
Subtleties in how individual legs move in LiDAR, or full body cues in 3D/camera measurements, are therefore lost. However, with end-to-end learning, since the data-driven model has access to the raw sensor input, it could theoretically infer actions based on such subtle information modalities. 

For end-to-end learning, having to train the entire perception component of a policy from sparse rewards can be inefficient.
An alternative is to use features from perception models which have been extensively trained with supervised training. This however, requires large quantities of labeled data.
Using unsupervised training, it is possible to learn rich encodings from unlabeled data.
These rich encodings can be shared across tasks.
In addition, these unsupervised models can be used to generate synthetic reproductions of the world, which can assist or replace simulation.
The question is, using current methods, can we train unsupervised representations which are practical for robot navigation?

We address this through the following contributions:
\begin{itemize}
\item We provide an open-source simulation environment, trained models, and useful tools for future end-to-end navigation benchmarking (see Fig.~\ref{fig:envs}).
\item We design several unsupervised learning architectures for the task of robot navigation among humans.
\item We compare existing approaches to each other and to ours, in seen and unseen environments, and demonstrate our approach on a real robot.

\end{itemize}



\section{Related Work}\label{sec:related_work}

Several state-of-the-art \ac{RL} methods for robot navigation make use of exact pedestrian positions in the input:
CADRL and following works \cite{chen2017decentralized, chen2017socially, everett2018motion} trained an \ac{RL}-based policy to avoid simulated agents with no static obstacles,
though real obstacles are handled out-of-policy in real testing.
%
SARL \cite{chen2019crowd} target the same domain as CADRL, with an architecture that focuses on pairwise interactions in order to model a useful value function. 
SOADRL \cite{liu2020soadrl} extends the SARL architecture to also take into accounts static obstacles, in some cases with a limited field of view. 

End-to-end methods avoid the need for detection and tracking of humans by using sensor data as input directly.
\citet{pfeiffer2017perception} explore using end-to-end supervised learning for robot navigation given 2D LiDAR measurements. Example trajectories are taken from an expert planner.
In \cite{pfeiffer2018reinforced} imitation learning is used, followed by end-to-end \ac{RL}, allowing for better performance and harder navigation problems.
Both \cite{pfeiffer2017perception, pfeiffer2018reinforced} have a static environment, with no dynamic agents.
CrowdMove \cite{fan2020distributed} shows a robot navigating in a crowded space using an end-to-end \ac{RL} policy. The learning agent is trained to solve a joint planning problem, where multiple instances of itself all simultaneously navigate past each other and obstacles to reach their goal. 

\citet{ha2018recurrent} raise the question of whether unsupervised learning can assist end-to-end \ac{RL}. Their world models approach is able to beat state-of-the-art performance on the CarRacing openAI gym environment by splitting the end-to-end task into an unsupervised latent feature learning task and a control policy learning from latent features \ac{RL} task.
A stated potential of this approach is the ability to use the unsupervised model to replace simulation, and they demonstrate a proof of concept for this idea on the Doom gym environment.
\citet{piergiovanni2018learning} applies this ``dream" idea to real-world robot navigation, in the limited scope of navigating to or avoiding an object based on camera images. 
%
GameGAN \cite{kim2020learning} pushes the neural simulation concept further by approximating the pac-man game through unsupervised learning of a \ac{GAN} architecture.
%
Recent work \cite{radford2018gpt, radford2019gpt2, brown2020gpt3}, also demonstrated that the unsupervised learning of sequences can deliver impressive results simply through increase in scale. They demonstrate state-of-the-art performance in various natural language processing tasks, using their approach. 







\section{Navigation Simulation Environment}

The NavRepSim environment is designed with \ac{RL} applications in mind. It aims to apply to any range-based sensor navigation problem. The goal of open-sourcing this simulator is to make it easier for anyone to reproduce state-of-the-art solutions for learning-based navigation.

The simulator offers 23 fixed maps as well as the possibility to procedurally generate maps containing random polygons (see Fig.~\ref{fig:envs} for examples of the available maps). The offered maps were designed to imitate those used in prior work. Note that pre-made gym environments are made available to reproduce the specific training domains of other paper. We also created several new maps from real LiDAR data. Overall, we provide environments that range from simplified and highly synthetic to more realistic maps containing real-world sensor noise.

In addition to having different maps, the learning environments can also vary in terms of the simulated robot(s), the simulated human agents as well as the learning curriculum. The simulated robot can be either holonomic or differential-drive. The inertial physics of the robot base can be simulated with high fidelity or simplified as instantaneous velocities. Further, collisions can either lead to damage (reflected as a negative reward) or instant episode termination. Users can also specify the number of robots in the environment, since joint training of several robots in the same simulation is also possible. The number and behavior of the simulated human agents can also be adjusted. Agents in the environment may be static or dynamic, they may be rendered as moving legs or ellipses, and they can be controlled using different policies (constant velocity, ORCA~\cite{van2008reciprocal}, global planning). Finally, we offer two learning curricula. The first uses a constant episode set-up, picking different maps and agent locations for every episode, while the second varies the environment difficulty (more human agents and more obstacles in the case of procedurally generated maps) based on the policy's success.

Care is taken to maintain simulation efficiency, with most environments running at more than 100 iterations per second in a single thread (on a laptop, Intel Core i7-6600U CPU @ 2.60GHz x 4). For comparison, the CarRacing gym environment used in \cite{ha2018recurrent} runs at the same speed on the same machine.

\begin{figure}[t]
    \centering
    \includegraphics[width=0.8\linewidth]{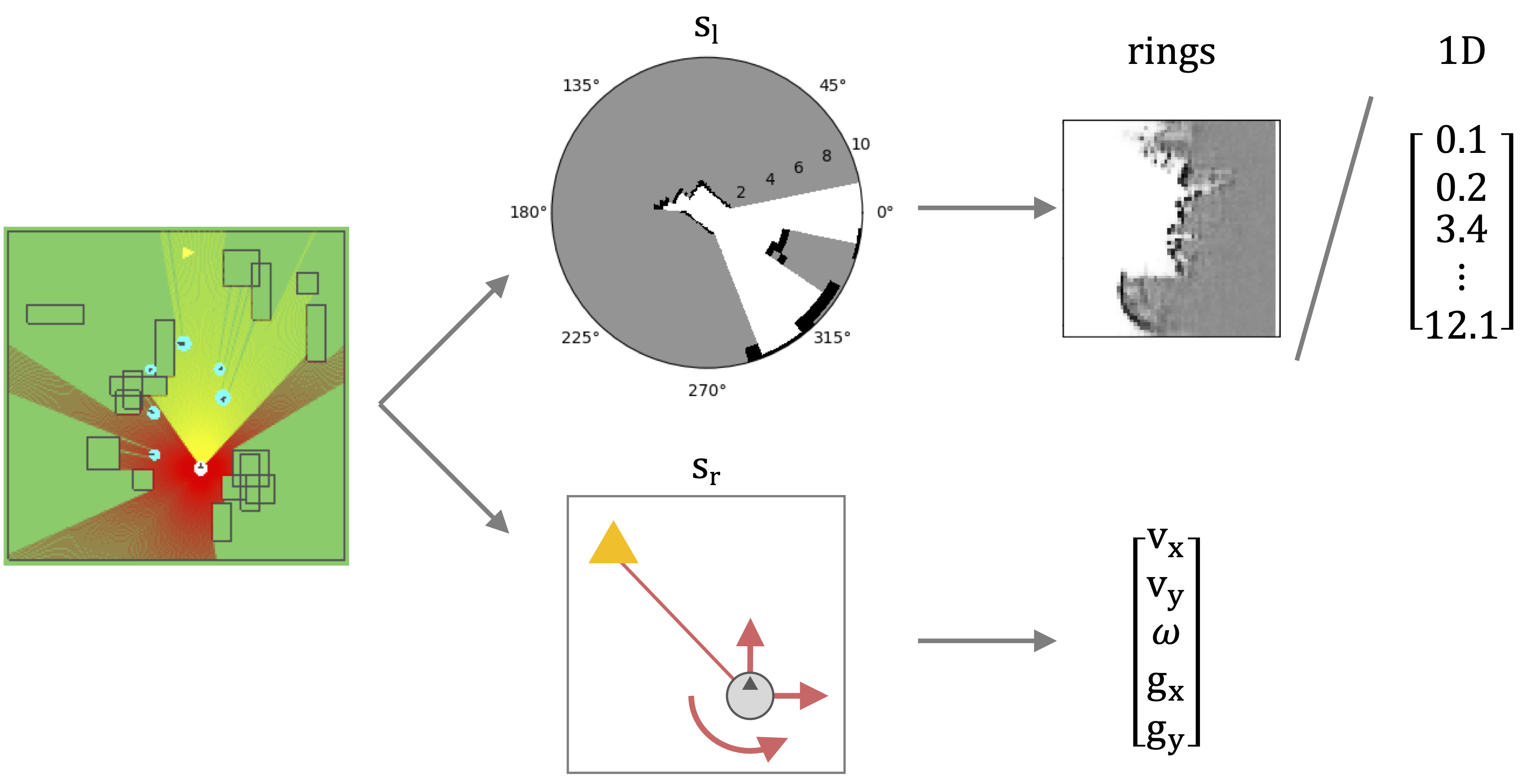}
	\caption{The state representation obtained from the world (left) combines both the LiDAR observation ($s_l$), and goal/velocity observation ($s_r$). The LiDAR observation is either represented as a $64 \times 64$ matrix of occupancy probability (rings) or a $1080$ vector of distances (1D).} 
	\label{fig:state}
\end{figure}

\section{Learning-based Navigation}\label{sec:learning_based_navigation}



A core purpose of this work is to study the possible variants of learning-based robot navigation. The basic setup common across many state-of-the-art methods is a ground robot equipped with a range sensor (often a 2D planar LiDAR). The state space used in the learning formulation is derived from the incoming range data. As discussed in Section~\ref{sec:related_work}, several works also assume that live pedestrian tracks are available to the robot, however in our work we do not make this assumption. The robot action space is continuous and is represented as a vector in $\mathbb{R}^3$, $\textbf{a} = \left[v_x, v_y, \omega\right]$. Each of $v_x$, $v_y$, $\omega \in [-1, 1]$ correspond to the robot's translational and rotational velocity in its base-link frame. Here we consider a holonomic robot, however the work can be extended to diff-drive robots by simply requiring $v_y=0$.

The scalar reward function for time-step $t$ is 
$r^t = r_s^t + r_c^t + r_d^t + r_p^t$.
The first term, $r_s^t$, is the success reward, which is 100 if the goal is reached and 0 otherwise. The collision reward, $r_c^t$, is -25 if the robot collides and 0 otherwise. The danger reward, $r_d^t$, is equal to -1 if the robot is closer than 0.2m to an obstacle or agent, and 0 otherwise. Finally,
$r_p^t$ is the progress reward, which is equal to the difference between the previous distance to goal and new distance to goal.

\subsection{State space variants}\label{sec:state_space_variants}
The state representation $s^t$ is a combination of the LiDAR observations $\boldsymbol{s_l^t}$ and the robot velocity and goal position in the robot frame $\boldsymbol{s_r^t}$ (see Fig.~\ref{fig:state}).
%
In this work, we consider two methods for representing LiDAR measurements and compare their effects when used as inputs to a learned navigation policy.
The first representation is simply a 1-dimensional vector of the 1080 raw distance values from the LiDAR normalized to $[0, 1]$. We refer to this as the \textbf{1D} representation.
We also consider a probabilistic representation for the input LiDAR data.
This representation maps the space around the robot as a polar grid ($\mathbb{R}^{64 \times 64}$), with evenly distributed discrete angular sections and exponentially distributed discrete radius intervals, leading to a greater resolution closer to the robot compared to regions that are farther away. Given the LiDAR scan, each grid cell is assigned a value $p \in \left\{0, 0.5, 1\right\}$ for being unoccupied, unknown or occupied, respectively.
We refer to this two-dimensional probabilistic LiDAR representation as the \textbf{rings} representation.
%
%

\subsection{NavRep: Unsupervised representations for navigation}
\begin{figure}
    \centering
    \includegraphics[width=0.6\linewidth]{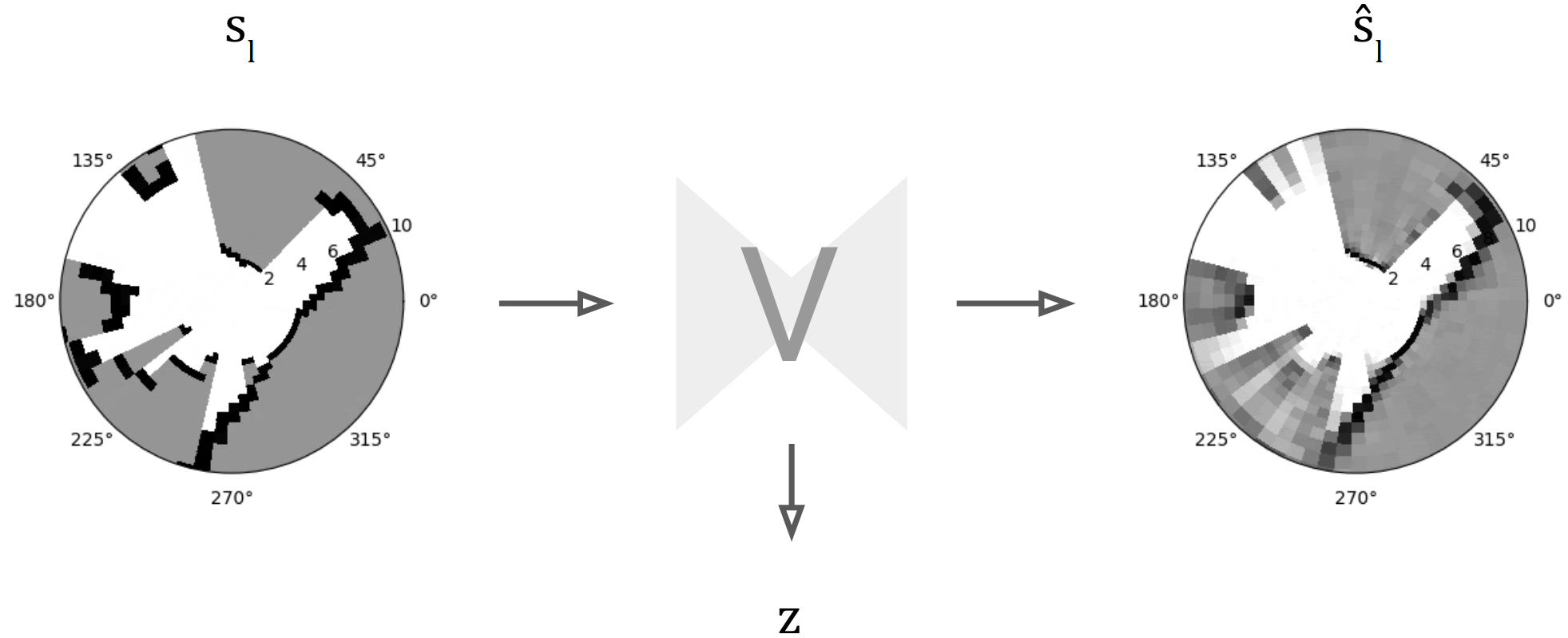}
	\caption{Visualization of the V auto-encoding module's inputs and outputs. Note that the LiDAR observation $s_l$ can be either in the rings or 1D representation.} 
	\label{fig:varch}
\end{figure}
\begin{figure}
    \centering
    \includegraphics[width=\linewidth]{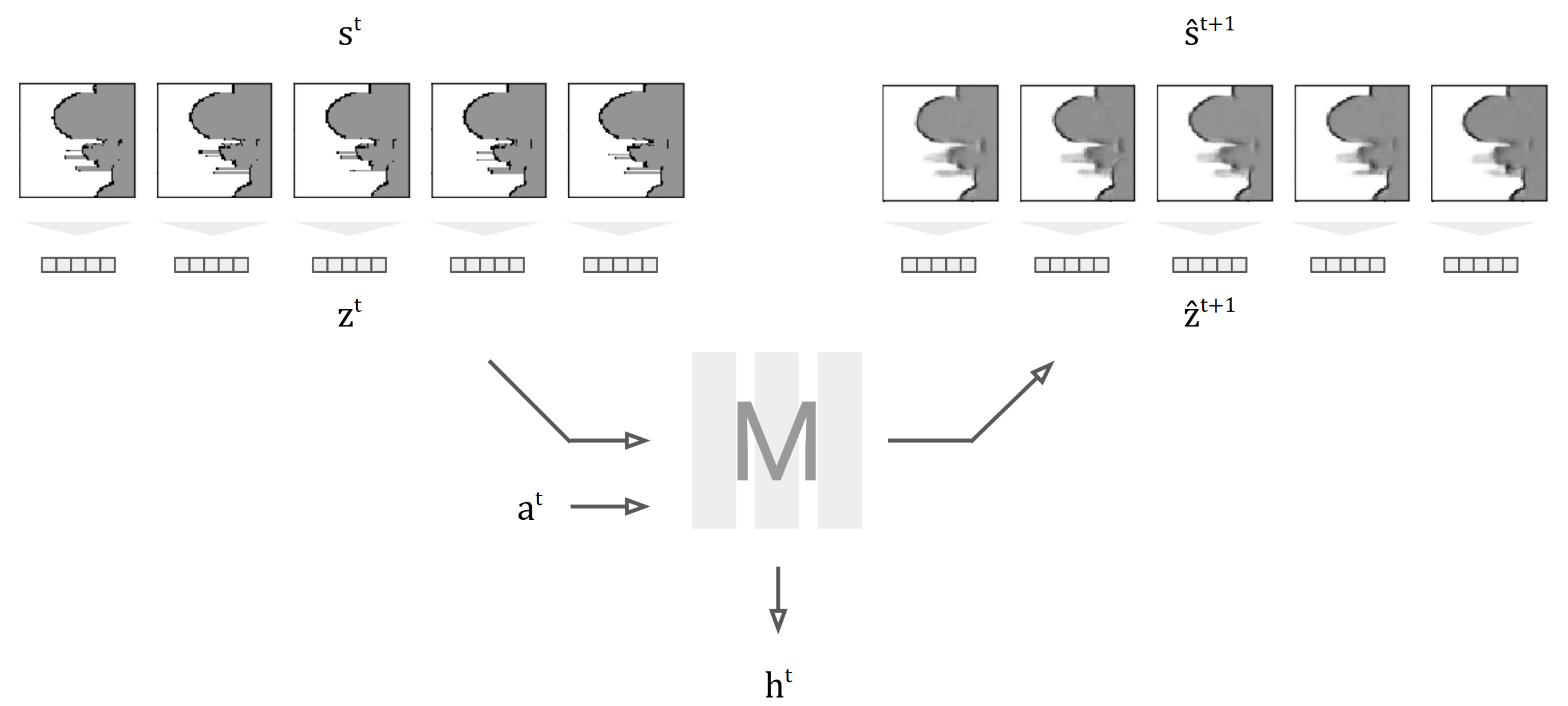}
	\caption{Visualization of the M prediction module's inputs and outputs. The $z$ latent features for each observation in the sequence are first encoded by the V module. These are passed to M along with the sequence of actions $a$. Each predicted $\hat{z}^{t+1}$ in the sequence is then decoded by V.} 
	\label{fig:march}
\end{figure}

In contrast to end-to-end learning approaches, NavRep exploits modular network architecture and training. Following the approach of~\cite{ha2018recurrent}, we make use of three main modules,
\begin{itemize}
    \item \textbf{V} (LiDAR encoding): a \ac{VAE} trained to reconstruct the LiDAR state,
    \item \textbf{M} (prediction): a \ac{LSTM} trained to predict future state sequences, and
    \item \textbf{C} (controller): a small \ac{FCN} (2-layer perceptron) trained to navigate using the latent representations of V and M as input.
\end{itemize}
For brevity, we refer to the latent encoding of V as $z$, and the latent encoding of M as $h$. These modules are illustrated Figs~\ref{fig:varch} and \ref{fig:march}.



\subsubsection{Transformer architecture}

From this basic set of modules, we offer our first NavRep learning variant. Rather than an \ac{LSTM} as in~\cite{ha2018recurrent}, we instead make use of the attention-based \textbf{Transformer} architecture~\cite{vaswani2017attention} to generate predictions of the future LiDAR encodings.

\subsubsection{Joint vs modular training}
Our second NavRep learning variant is designed to provide insight into the effect of training V and M jointly or separately. In~\cite{ha2018recurrent}, the authors first train V to minimize reconstruction loss before training M to minimize the $z$ prediction loss while keeping the V network weights fixed. Here we compare their \textbf{modular} training regime with one that trains V and M \textbf{jointly}.

\subsubsection{Input features to C}
\begin{figure}[t]
    \centering
    \includegraphics[width=\linewidth]{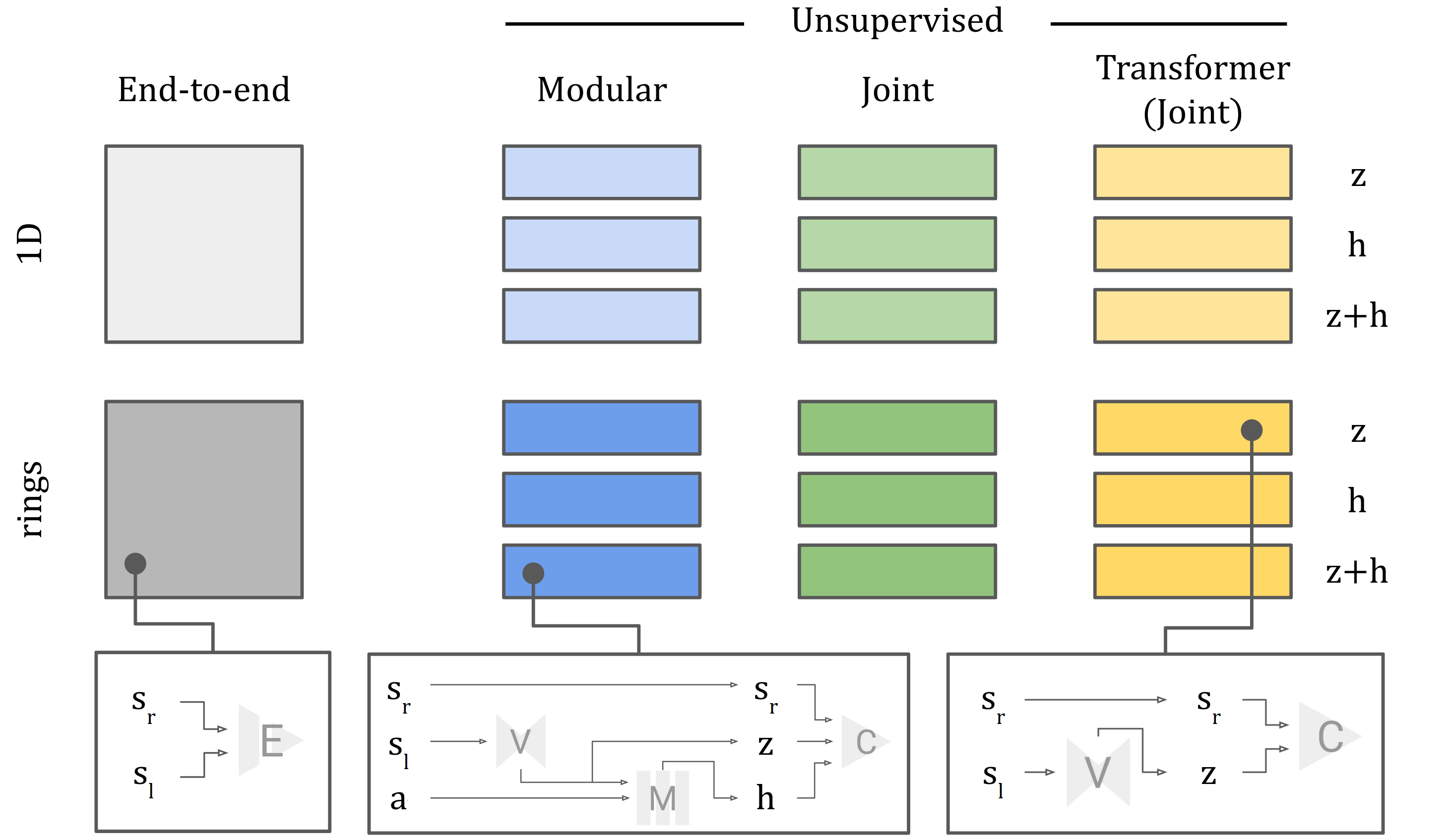}
	\caption{Overview of the architectures trained in this work. For all approaches there are 2 LiDAR input variants. In addition, the 3 unsupervised-learning-based approaches each have 3 latent feature variants. Simplified schema of the components are detailed for 3 abitrarily selected architectures.} 
	\label{fig:archoverview}
\end{figure}

The final set of NavRep variants that we consider are the inputs used by the controller to compute the robot action. We compare between using $z$, $h$ or $\left[z,h\right]$, where the latter was proposed in~\cite{ha2018recurrent}. An overview of all the learning architecture and training variants that we study is provided in Fig.~\ref{fig:archoverview}.

\subsection{Network implementation and training}

\subsubsection{Hyperparameters}

All V modules are composed of a 4-layer \ac{CNN} encoder, followed by 2-layer \ac{FCN}, and 4-layer \ac{CNN} decoding blocks. $z$ latent features are of size 32, $h$ latent features are of size 64.
The LSTM-based M module has 512-cells. The Transformer-based M module is composed of 8 causal-self-attention blocks with 8 attention heads each. The modular V and M modules have respectively 4.3 million and 1.3 million trainable parameters. The V+M transformer model has 4.8 million trainable parameters.
More details available in the open-source implementation.

\subsubsection{V}
was trained using minibatch gradient descent, with a typical VAE loss,
$$ L_{V} = L_{rec} + KL(z^t), $$
where $KL(z^t)$ is the Kullblack-Leibler divergence between the latent feature distribution parameterized by the encoder and the prior latent feature distribution. The reconstruction loss $L_{rec}$ is the mean squared error between input and output,
\begin{equation}\label{eq:reconstruction_loss}
L_{rec} = H(\hat{s}^t, s^t),
\end{equation}
where $H$ denotes the binary cross entropy, $s^t$ denotes the observed state (LiDAR representation) at time $t$ in the sequence.

\subsubsection{M}
was trained on batches of sequences. We define the loss function as the latent prediction loss,
\begin{equation}\label{eq:pred_loss}
   L_{M} = H(\hat{z}^{t+1}, z^{t+1}) + H(\hat{s_r}^{t+1}, s_r^{t+1}). 
\end{equation}
This adds the binary cross entropy loss between the predicted ($\hat{z}^{t+1}$) and the true ($z^{t+1}$) latent features to that of the predicted ($\hat{s_r}^{t+1}$) and the true ($s_r^{t+1}$) goal-velocity states.

\subsubsection{Joint V+M}
was trained with a loss function composed of the sum of the reconstruction and prediction losses from \eqref{eq:reconstruction_loss} and \eqref{eq:pred_loss}. However in this case, for the rings variant the prediction loss is
$$ L_{V+M} = H(\hat{s}^{t+1}, s^{t+1}), $$
the binary cross entropy loss between the reconstructed state prediction $\hat{s}^{t+1} = V(\hat{z}^{t+1})$ and true next state $s^{t+1}$. For the 1D LiDAR variant we used the mean squared error instead of the cross-entropy loss as the predictions are not distributions, but direct values.

\subsubsection{C}
was trained using the \ac{PPO} algorithm \cite{schulman2017proximal, hill2018stable}. Each C module was trained up to 3 times with varying random seeds in order to mitigate initialization sensitivity in the results. Training was stopped after 60 Million training steps.

\subsubsection{Training data}

The V and M modules were trained with data extracted from the same environment in which the C module was later trained. 
To generate training sequences, the ORCA~\cite{van2008reciprocal} policy was used to control the robot, and the number of agents in the scene was increased to ensure sufficient observations of dynamic objects and prevent imbalance in the dataset. A simple model of leg movement was used to render dynamic agents as moving legs in the generated LiDAR data and a time-step of 0.2s was used for updating the agents' positions.

The C module itself was trained in the SOADRL-like environment (shown in Fig.~\ref{fig:envs}), with randomly generated polygons and agents (every episode is unique). A bounding obstacle had to be added to prevent the policy from learning a simple risk-averse heuristic which consists in going around the entire area. During training, actions for the C module were limited to x-y movement, with 0 rotation. However, the robot's initial angular position was randomized for each episode.

Curriculum learning was used to improve the training of the policy: every failed episode lead to a lower number of polygon obstacles and agents, conversely, every successful episode lead to an increase. The maximum number of agents and polygons was capped to 5 agents and 10 obstacles.

\subsection{End-to-end learning baselines}\label{sec:e2e_learning}
As a baseline, we implement a typical end-to-end \ac{RL} approach. This approach attempts to learn a navigation policy $\pi(a, s)$ predicting action probabilities directly from LiDAR inputs.

Two variants of end-to-end learning are implemented: the first taking the 1-dimensional LiDAR representation as the state input, and the second taking 2-dimensional rings representation as the state input. For the value and policy model, we use a \ac{CNN} for feature extraction, followed by fully connected layers. To make comparisons fair, the same \ac{CNN} architecture and dimensions are used as in the V and C models described in the following subsection.

\section{Experiments and Results}

\subsection{Worldmodel error for unsupervised learning architectures}
\begin{figure}
    \centering
    \includegraphics[width=\linewidth]{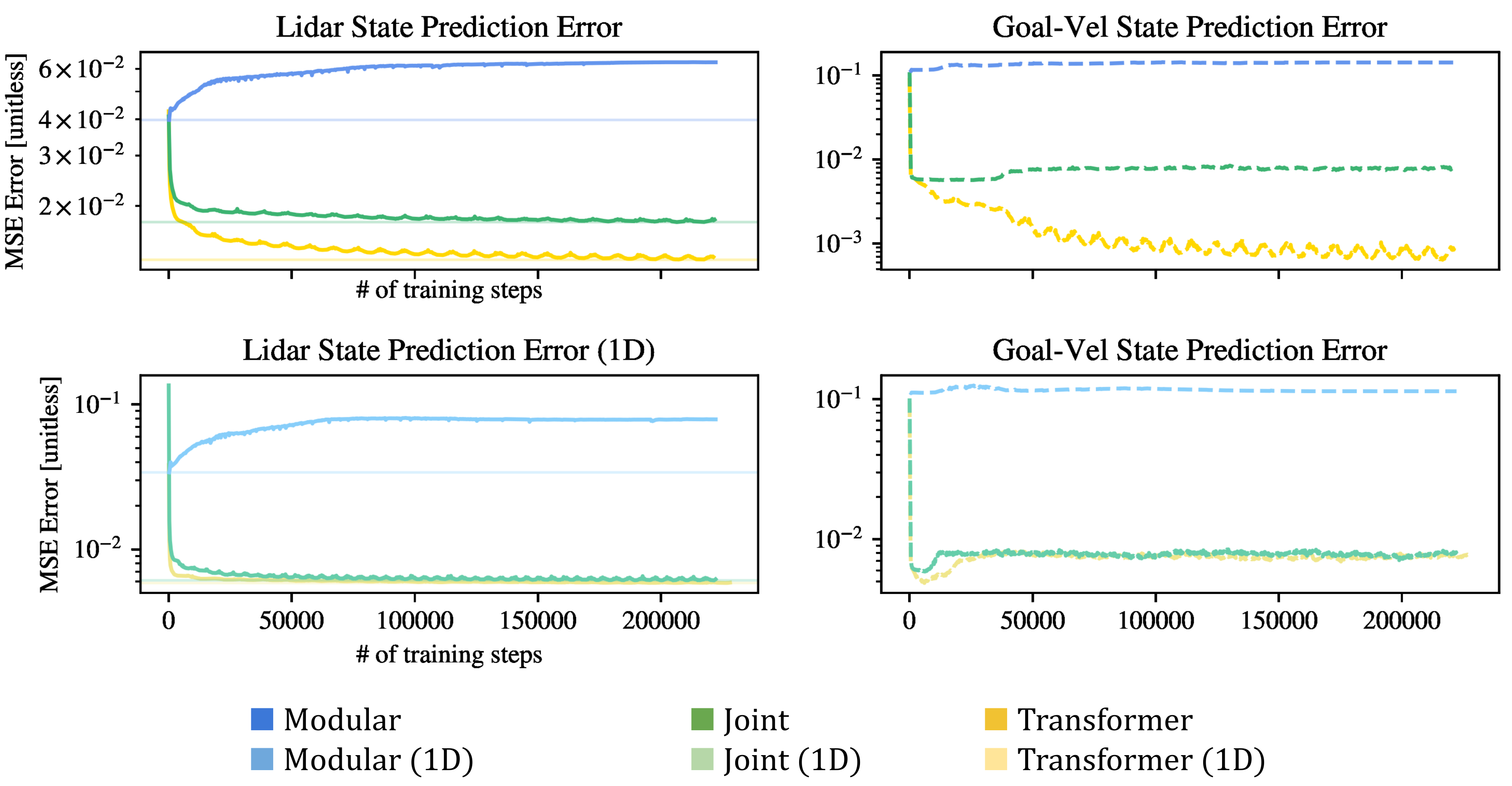}
	\caption{Comparison of the \textit{worldmodel error} for the joint, modular and transformer architectures, for their (top) rings LiDAR state representation variant, and ((bottom) 1D variant. This error is calculated and shown separately for the LiDAR state $s_l$ (left) and goal-velocity state $s_r$ (right).} 
	\label{fig:w_error}
\end{figure}

The six joint, modular and transformer architectures (Fig.~\ref{fig:archoverview}) have differing loss functions, which makes it difficult to compare their losses during training. 
To this end, we define the ``worldmodel error", a next-step prediction error term that allows meaningful comparison across architectures,
$$ E_{\textit{worldmodel}} = MSE(\hat{s}^{t+1}, s^{t+1}). $$
However, this error function does not produce comparable results for the rings vs. 1d LiDAR inputs, as they do not have the same dimensions and scale. Instead, these comparisons are shown separately in Fig.~\ref{fig:w_error}. The worldmodel errors shown here are calculated for the training sets, which compares each architecture's ability to learn in an unsupervised manner.
It can also be understood as the accuracy of 'Dreams' generated by each architecture.

\subsection{Controller training performance} 
\begin{figure}
    \centering
    \includegraphics[width=\linewidth]{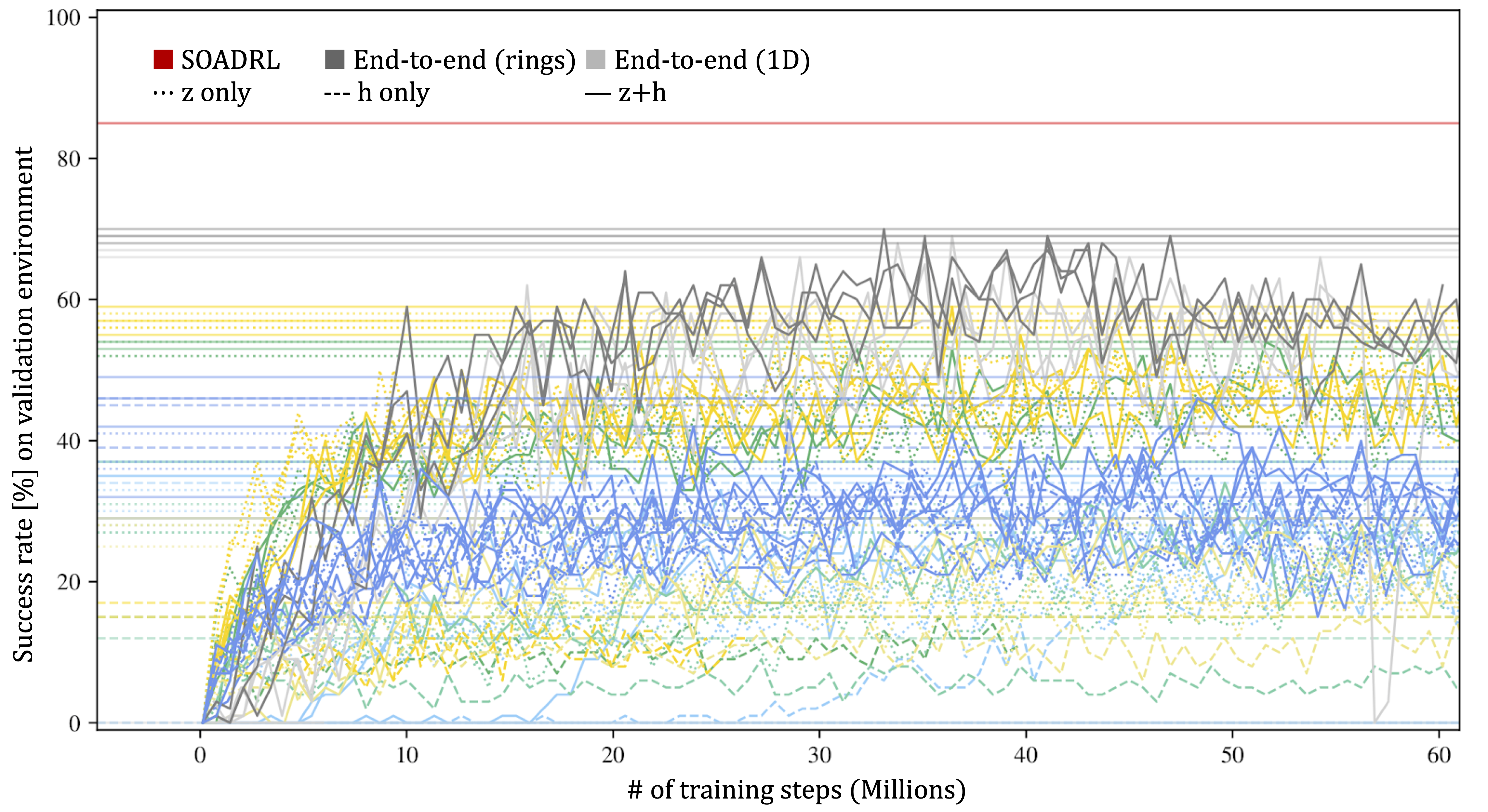}
	\caption{3 training sessions were performed for each of the two end-to-end and 18 NavRep architectures to account for the influence of network initialization on the results (legend as in Fig.~\ref{fig:w_error}). Horizontal lines mark the maximum success rate of each curve. The SOADRL success rate allows for comparison with methods that use exact human locations as input.} 
	\label{fig:training}
\end{figure}

Though the training environment features curriculum learning through adaptive difficulty, a separate validation environment with fixed difficulty is used to quantify the absolute progress of each model every 100'000 training steps. This validation is similar to the training environment, except that it remains on the maximum difficulty, i.e. maximum number of agents and obstacles, and that it consists of 100 episodes with arbitrary but deterministic initializations. Success rates (number of validation runs in which the robot reaches the goal without collision) are shown in Fig.~\ref{fig:training}.
\subsubsection{Discussion}
Looking at the modular, joint and transformer models, it seems that the worldmodel error and learned controller performance are correlated. This suggests that better reconstruction and prediction performance in the V and M modules leads to latent features which are more useful for the control task. Nevertheless, it can also be seen that the end-to-end approaches perform better in the validation tasks than the NavRep controllers.
Finally, none of the models are able to surpass the performance of SOADRL, which has access to exact positions of pedestrians. 
%

\subsection{Test performance}
Each model was tested in 27 unseen scenarios (3 maps, 9 scenarios each). At least 300 episodes were run in each scenario, for a total of 10'000 test episodes per model. These scenarios are designed to represent a wide range of possible real-life situations that can help to identify specific issues with the learned controller, such as global planning, agent avoidance and crowd aversion. For each map, the first 6 scenarios are taken from~\cite{dugas2020ian}, and the remaining 3 are
(7) a trivial scenario where the robot is close to the goal with no obstacle in between,
(8) an easy scenario with a close goal, and a single obstacle between robot and goal, and
(9) a typical dynamic avoidance task where all agents are placed on a small circle with goals on the opposite side and no obstacles in between. The scenario definitions and selection was fixed before any testing took place to avoid bias.


\begin{figure}
    \centering
    \includegraphics[width=\linewidth]{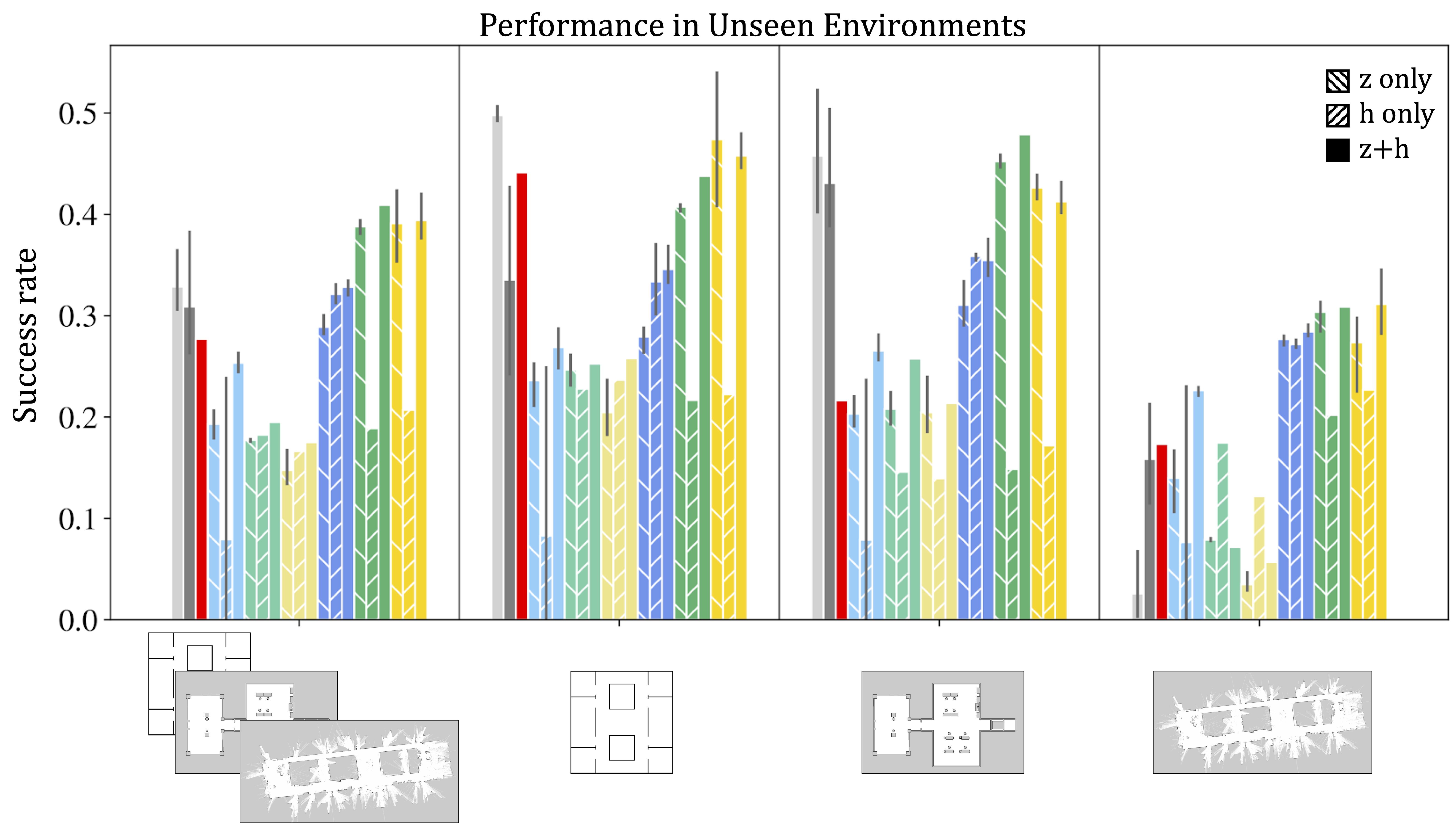}
	\caption{Mean performance of each model in the testing environments. Legend as in Figs~\ref{fig:w_error}, \ref{fig:training}, with error bars denoting the minimum and maximum score across random seeds. Overall performance is shown (left), as well as per-map performance from the simple, complex, and realistic maps.} 
	\label{fig:testing}
\end{figure}
\begin{figure}
    \centering
    \includegraphics[width=\linewidth]{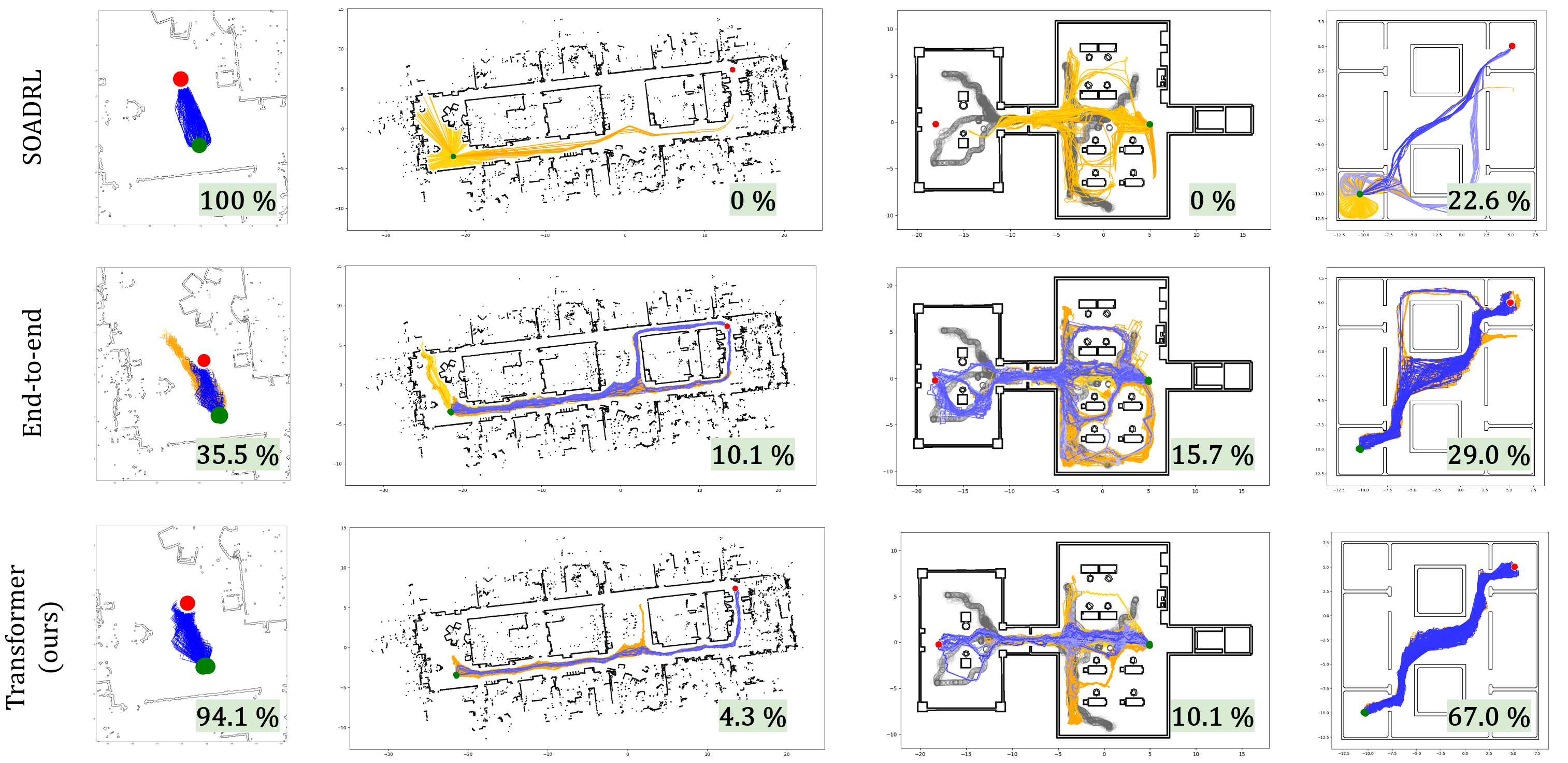}
	\caption{Comparison of trajectories executed by the best performing models during testing in unseen environment. From left to right, the scenarios are 1) simplest scenario in realistic map, 2) global planning task in realistic map, 3) full planning task (global, static, dynamic avoidance) in complex map, 4) global planning in simple map. Orange trajectories are trajectories which resulted in failure to reach the goal, through collision or time-out. Blue trajectories are successful trajectories. Grey circles indicate the positions of dynamic agents. Overlaid is the percentage of successful trajectories.} 
	\label{fig:traj}
\end{figure}


\subsubsection{Discussion}
In general it can be seen that the end-to-end models perform suffer a greater drop in performance from training to testing, whereas the NavRep models have more consistent performance. In particular, the 1D LiDAR end-to-end model fails completely in the realistic map even though it is a top performer during training. In addition, when investigating the low performance of SOADRL, we saw that most failures were due to crashes into obstacles and not agents.

In the same test environment, a traditional planner based on the \ac{TEB} approach from the widely-used ROS navigation stack achieves 76\% success rate.

The low success rates of all models on long-range navigation scenarios (Fig.~\ref{fig:traj}) indicates that global navigation is a challenge for RL-based methods. We also see that several works \cite{chen2017decentralized, liu2020soadrl, everett2018motion} avoid this problem by dealing with local navigation only. 
Looking at other failures, we see that in several cases, the end-to-end and unsupervised models don't learn to proceed slowly in the presence of ``danger'' (other dynamic agents). Instead they move as fast as possible, or are stuck when available space becomes narrow. This can perhaps be improved through reward design.
Finally, in the challenging dense static crowd scenario, though most attempts fail, a few are able to succeed by avoiding the dense area, which implies taking a significantly longer route. As shown in~\cite{dugas2020ian}, this is a challenging situation even for traditional planners. 

\subsection{Analysis of learning variants}\label{sec:analysis}
\subsubsection{Impact of rings}
Rings encodes the space around the robot as values between 0 and 1 in a pseudo-probabilistic representation of occupancy. It seems that allowing the NavRep model to infer from and predict occupancy uncertainties explicitly makes its task easier than inferring directly from the raw LiDAR scan.
In addition, rings provides a higher-level abstraction for the CNN to operate in, meaning that it requires less depth, while scaling the spatial resolution information to focus on regions close to the robot.
This is in line with how sensors typically work and also matches the way that, in obstacle avoidance, more attention should usually be paid to close objects.
The difference in resolution sensitivity between the 1D and rings representations could also explain the dramatic performance drop observed in the realistic map for the end-to-end model. Local features in the LiDAR representations for the simple vs. realistic maps vary more in the 1D case than in the rings. Moreover, rings more easily allows using the same model with different sensors of varying resolution.
%

\subsubsection{Joint vs modular training}
Previous work \cite{ha2018recurrent} recommends modular training, that is, training V, M and C separately, and to use both $z$ and $h$ as inputs to the C module. However, in this work we find that jointly trained V+M models provide superior performance in almost all tasks. The computational cost of joint training is furthermore not significantly greater than for modular training, though memory usage does increase due to having to store sequences of rich observations rather than compressed latent features.

\subsubsection{Latent feature usefulness}
When looking at model performances both in seen and unseen environments, our results do not support the idea that using both $z$ and $h$ latent features is advantageous. Though in some cases models trained with $z+h$ features performed slightly better, this improvement is not significant when compared to the variance between random seeds. In our experience, the performance cost of including $h$ latent features in the C training regime is great: training a $z+h$ C model takes approximately 9 times longer than for an equivalent $z$-only C model. $h$-only models have it worst, with low test performance and slow training times.

\subsection{Real-robot trials}
\begin{figure}
    \centering
    \includegraphics[width=\linewidth]{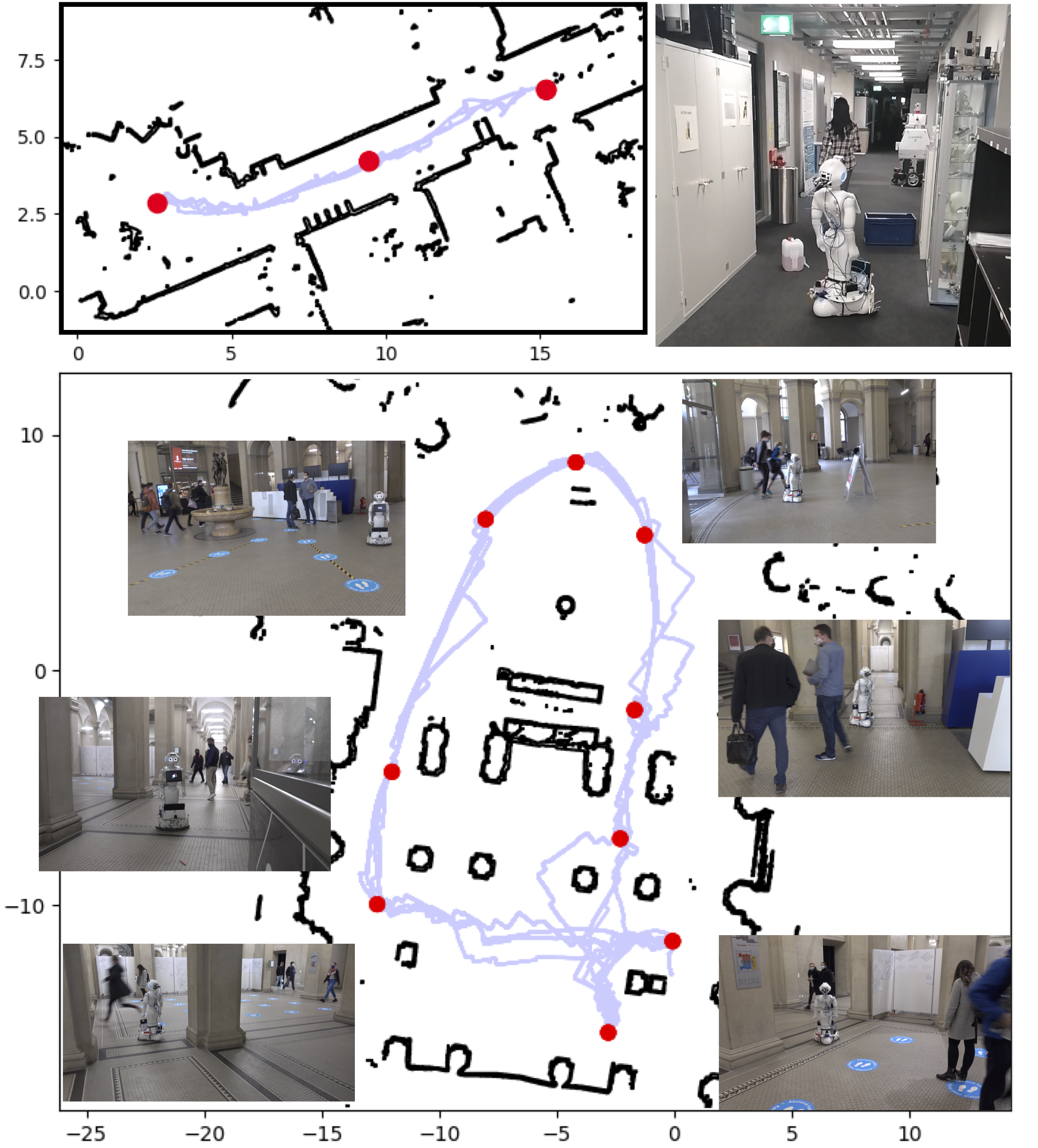}
	\caption{We test the proposed method on a real robot, where the learned policy is repeatedly able to avoid obstacles and people and reach its goals. Goals are shown in red, successful trajectories in blue, images taken during the experiment are displayed next to the locations they were taken in. Each sequence was run three times in a row.} 
	\label{fig:hg2}
\end{figure}

We took our Transformer architecture (rings, $z$-only) and implemented it on a Pepper robot (Fig.~\ref{fig:hg2}). Our robot has a quasi-holonomic base and two 270$^\circ$ LiDARs merged together to form a 1080-vector combined range scan. The robot was given several waypoints in a space mixed with people and obstacles, and tasked to patrol between waypoints. 

The learned controller achieved 100\% success in the real environment, reaching its goal for all trajectories. However, we observed issues which show a need for improvement.
(i) Smoothness: the commands selected by the controller are aggressive, leading to jerky motion.
(ii) Reluctance to go into tight spaces: when faced with narrow but traversable areas, the controller oscillates around the entrance but eventually makes it through.
(iii) Getting stuck: we observed the planner spend several seconds oscillating in front of an obstacle before finally passing on one side.

These issues appear to be in part due to the following causes:
(i) Inertia in the dynamics of the robot is not modelled in the training environment, which causes sub-optimal control in the policy.
(ii) Sensor noise is not present in the simulated environment. As a result, when it occurs in the real scenario, the controller can react unpredictably.
We surmise that there are further unknown reasons. Such ``unknown unknown'' effects could possibly be addressed via continual learning by refining the model in the real world.

\section{Conclusion}

Training robust sensor-to-control policies for robot navigation remains a challenging task. NavRep methods do not dominate over end-to-end methods, but we do see interesting properties and trade-offs between the two. Though less suited to learning specific tasks, unsupervised representations display several benefits, such as more consistent performance across general tasks, modularity, and the potential to generate ``Dream" environments~\cite{ha2018recurrent}, which could assist or replace simulation. The popularity of these methods is additionally propelled by the continued increase in compute and memory availability. Moving forward we expect to see larger unsupervised models with richer modalities, that create more convincing approximations of reality and provide more useful and discerning features.

Nevertheless, navigation-RL still has a long way to go, should one aim to replace traditional planners. Among the current issues are global planning, sim-2-real constraints, and safety. Potential avenues include more rigorous exploration of reward functions, more realistic training environments, and continual learning. Given that standardized gym-like environments could help harmonize future research, we hope that by making NavRep and NavRepSim publicly available and easy to use, progress in this domain can be accelerated and to improve collaboration between researchers.

\addtolength{\textheight}{-12cm}


\bibliographystyle{IEEEtranN}
\footnotesize
\bibliography{references}

\end{document}